\documentclass{article} % For LaTeX2e
\usepackage{colm2024_conference}

\usepackage{microtype}
\usepackage{hyperref}
\usepackage{url}
\usepackage{booktabs}
\usepackage{cuted}   % For the strip environment
\usepackage{afterpage}
\usepackage{xcolor,colortbl}
\usepackage{mdframed}
\usepackage{amssymb}
\usepackage{listings}
\usepackage{graphicx}
\usepackage{subcaption}
\usepackage{enumitem}
\usepackage{verbatim}
\usepackage{caption}
\usepackage{amsfonts}       % blackboard math symbols
\usepackage{amsmath}
\usepackage{nicefrac}       % compact symbols for 1/2, etc.
\usepackage{pifont}

\colmfinalcopy

\newmdenv[
  backgroundcolor=gray!20, % Set background color to light gray
  linewidth=0pt,            % No border
  innerleftmargin=10pt,     % Padding inside the frame
  innerrightmargin=10pt,
  innertopmargin=10pt,
  innerbottommargin=10pt
]{customquote}

\newcommand{\nw}[1]{}

\definecolor{kellygreen}{rgb}{0.3, 0.73, 0.09}
\definecolor{alizarin}{rgb}{0.82, 0.1, 0.26}

\newcommand{\cmark}{{\color{kellygreen} \ding{51}}}
\newcommand{\xmark}{{\color{alizarin} \ding{55}}}

\newcommand{\ourmethod}{UISim}

\usepackage{authblk}

\title{\ourmethod: An Interactive Image-Based UI Simulator for Dynamic Mobile Environments}

\author[1, $\spadesuit$]{Jiannan Xiang}
\author[2]{Yun Zhu}
\author[2, $\spadesuit$]{Lei Shu}
\author[2]{Maria Wang}
\author[2]{Lijun Yu}
\author[2]{Gabriel Barcik}
\author[2]{James Lyon}
\author[2]{Srinivas Sunkara}
\author[2]{Jindong Chen}

\affil[1]{University of California, San Diego (UCSD)}
\affil[2]{Google DeepMind}
\affil[$\spadesuit$]{Correspondence to: \{\text{jixiang@ucsd.edu}, \text{leishu@google.com}\}.}

\begin{document}

\maketitle

\renewcommand{\thefootnote}{\fnsymbol{footnote}}
\footnotetext[2]{A demonstration video is available at \url{https://youtu.be/QZOhjFQRZxs}.}

\begin{abstract}
  Developing and testing user interfaces (UIs) and training AI agents to interact with them are challenging due to the dynamic and diverse nature of real-world mobile environments. Existing methods often rely on cumbersome physical devices or limited static analysis of screenshots, which hinders scalable testing and the development of intelligent UI agents. We introduce \ourmethod, a novel image-based UI simulator that offers a dynamic and interactive platform for exploring mobile phone environments purely from screen images. Our system employs a two-stage method: given an initial phone screen image and a user action, it first predicts the abstract layout of the next UI state, then synthesizes a new, visually consistent image based on this predicted layout. This approach enables the realistic simulation of UI transitions. \ourmethod\ provides immediate practical benefits for UI testing, rapid prototyping, and synthetic data generation. Furthermore, its interactive capabilities pave the way for advanced applications, such as UI navigation task planning for AI agents. Our experimental results show that \ourmethod\ outperforms end-to-end UI generation baselines in generating realistic and coherent subsequent UI states, highlighting its fidelity and potential to streamline UI development and enhance AI agent training.
\end{abstract}

\section{Introduction}

Developing and testing user interfaces (UIs), particularly for mobile devices, is a complex and often resource-intensive process. Traditional methods rely heavily on manual screen captures and scripted interactions performed on physical devices or emulators~\citep{stone2005user,sridevi2014user,thimbleby1990user,carroll1988interface}. These methods are time-consuming, inflexible and do not scale well for comprehensive testing or agent training. Real-world UI environments are inherently dynamic and diverse, presenting significant challenges for automated analysis, rapid prototyping, and the development of robust UI navigation agents. As such, the ability to efficiently simulate these environments, allowing for interactive exploration and generation of plausible next states, is crucial for advancing both UI development and AI-driven human-computer interaction.

\begin{table}[t]
\centering
\caption{A comparative analysis of UI simulation and generation paradigms. This table evaluates various methods against critical features including dynamic state simulation, the ability to exert fine-grained structural control, and overall scalability. UISim (Ours) is demonstrated to address the combined challenges that other approaches face, offering a comprehensive solution for interactive mobile UI environments.}
\small
\resizebox{\textwidth}{!}{
\begin{tabular}{lccc}
\toprule
Methods                       & Dynamic State Simulation & Fine-Grained Structural Control & Scability \\ \midrule
Physical Emulator             & \cmark                        & \cmark                               & \xmark                       \\
High-Level UI Spec Generation & \xmark                        & \cmark                               & \cmark                       \\
Static Screenshot Analysis    & \xmark                        & \xmark                               & \cmark                       \\
End-to-End Image Generation   & \cmark                        & \xmark                               & \cmark                       \\
\ourmethod\ (\textbf{Ours})     & \cmark                        & \cmark                               & \cmark                       \\ \bottomrule
\end{tabular}
}\label{tab:compare}
\end{table}

Existing approaches to UI simulation and automation often fall short in critical areas, as shown in Table \ref{tab:compare}. Some methods focus on generating UIs from high-level specifications~\citep{usman2014model,usman2020automated,son2013mof,li2022uldgnn,chen2024egfe}, which limits their ability to accurately replicate the visual nuances of real-world  applications. Others rely on static screen captures~\citep{wu2021screen,beltramelli2018pix2code}, preventing dynamic interaction and exploration of subsequent states. While recent advances in image generation~\citep{podell2023sdxl,betker2023improving} offer promising directions, end-to-end approaches often struggle with the precise control needed for coherent UI state transitions and often produce less realistic visual output. An interactive, image-based simulator that can dynamically respond to user actions and generate high-fidelity, plausible next-UI screens remains a significant challenge, especially for complex real-world UIs. Such a system is crucial for enabling rapid prototyping, scalable testing of UI designs, and critically, for fostering the development of intelligent agents that can learn to navigate complex digital environments through rich visual feedback.

To address these pressing limitations, we introduce \ourmethod, a novel image-based UI simulator that enables dynamic and interactive simulation of mobile phone environments purely from screen images. Our system uniquely provides a flexible and scalable platform for exploring UI states and actions without requiring direct access to a physical device or a complex rendering engine. Given an initial phone screen image and a user action (e.g., open the email app), \ourmethod\ employs a robust two-stage method to generate the subsequent UI state: it first predicts abstract layout information describing the possible next screen based on the input action, then synthesizes a new, visually consistent UI image based on this predicted layout. The decoupled design breaks down the complex problem of image-to-image UI transformation into more manageable sub-problems, offering fine-grained control  over the content and structure of the simulated UI, inherently leading to higher fidelity and more diverse generation capabilities compared to end-to-end image generation methods. Our experimental results show that \ourmethod\ outperforms baselines by 36.73 on Fréchet Inception Distance, demonstrating its superiority in generating realistic and coherent subsequent UI states.

\ourmethod\ delivers immediate and substantial practical benefits across multiple domains. For developers and designers, it enables rapid prototyping and UI testing, allowing quick iteration on designs, visualization of user flows, and observation of application behavior under various simulated user interactions, drastically reducing the need for cumbersome manual device testing. For researchers, it facilitates the generation of vast amounts of synthetic UI data, crucial for training data-hungry machine learning models for UI analysis and automation, and evaluating UI automation in a reproducible environment. Furthermore, the simulator's inherent capability to generate plausible next UI states based on arbitrary user actions is important for advanced applications such as UI navigation task planning. By providing a "look-ahead" mechanism for AI agents to explore potential interaction sequences and their visual outcomes, \ourmethod\ can facilitate the development of more intelligent and adaptable agents capable of achieving complex, high-level goals like "send an email" or "book a flight" across diverse and unfamiliar applications. We believe this interactive, image-based simulator represents a crucial step towards more efficient and scalable development of both UI-centric applications and general-purpose AI agents interacting with human-designed interfaces.

\section{Related Work}

\paragraph{UI Simulation and Automation} User interface (UI) development and testing have traditionally relied on either manual procedures or tooling that requires direct access to application source code or runtime environments. Tools like Appium~\citep{appium}, UIAutomator~\citep{uiautomator}, and the Android Studio Emulator~\citep{android_emulator} support scripted interaction and visual inspection but are limited by their dependency on physical or emulated devices and their inability to simulate arbitrary UI states at scale. To address some of these limitations, model-driven UI generation approaches such as Model-Based Testing of GUI Applications and Automated UI Generation from Task Models~\citep{usman2014model,usman2020automated,son2013mof,li2022uldgnn,chen2024egfe} use high-level specifications to produce static UI layouts. However, these methods focus on synthesis rather than simulation and lack support for visually grounded state transitions or dynamic user interaction.

Other recent works aim to understand or reconstruct UIs from screen images. For example, \cite{beltramelli2018pix2code} and \cite{wu2021screen} attempt to infer UI structures from visual input, enabling applications like code generation or layout understanding. Nevertheless, these approaches are typically static—they do not model the dynamic behavior of UIs or generate plausible next states in response to user actions. Even simulation environments such as Rico~\citep{deka2017rico} focus more on data collection and replay, rather than enabling new state generation.

In contrast to these prior methods, our work introduces \ourmethod, an image-based UI simulator capable of predicting and synthesizing the next UI state purely from screen images and user actions. This allows for dynamic, visual, and interactive UI simulation without reliance on application internals or emulators. By decoupling structure prediction from image generation, our approach can flexibly and realistically model transitions across diverse UI environments, bridging a critical gap in UI automation and interaction research.

\paragraph{Image and Video Generation for Structured Interfaces} Recent years have witnessed the rapid progress of image~\citep{podell2023sdxl,betker2023improving,flux2024,saharia2022photorealistic} and video diffusion models~\citep{videoworldsimulators2024,veo,polyak2410movie}. These models excel at producing high-fidelity outputs from textual or visual prompts, and have demonstrated impressive generalization across natural scenes and object categories. However, applying them directly to user interface (UI) generation remains challenging. Unlike natural images, UI screens follow strict structural rules—buttons, text fields, and layout hierarchies must remain consistent and functional across transitions. General-purpose image generators often lack the spatial control and semantic consistency needed to model these structured visual domains.

Several works attempt to bring structure-awareness into generative models. For example, LayoutDiffusion~\citep{zheng2023layoutdiffusion} and ControlNet~\citep{zhang2023adding} incorporate layout or edge-map conditioning to guide generation, which is useful for structured scenes but not specifically tuned for interactive environments like mobile UIs. End-to-end image/video generation models often struggle with the inherent reasoning required for accurate next-screen prediction in UI environments. Unlike natural image generation, where the prompt directly describes the desired visual output, predicting a UI transition necessitates inferring how user actions logically alter the screen's layout, content, and functionality. General-purpose generative models typically lack the explicit mechanism for this kind of high-level semantic understanding and structural planning. Their one-to-one mapping from input to output, without an intermediate reasoning step, can lead to generated UI states that are visually plausible but structurally inconsistent or semantically incorrect with respect to the intended user interaction.

\ourmethod\ addresses this gap by introducing a structured two-stage pipeline specifically designed for UI simulation. It first predicts a layout conditioned on the current screen and user action, then synthesizes a high-resolution UI image from this layout. This decomposition enables fine-grained control over structure and visual fidelity, allowing our method to produce coherent and realistic transitions between states—capabilities that end-to-end visual generation models typically struggle to achieve in structured domains like mobile UIs.

\paragraph{UI Interactive Agents} Vision-based AI agents for UI navigation and automation are increasingly studied as an alternative to traditional rule-based systems, particularly for applications involving accessibility, automation, and testing. Early works such as DroidBot~\citep{li2017droidbot} explore Android UIs by building runtime state models from screenshots and UI hierarchies to guide interaction testing. More recent approaches have moved toward instruction-following agents that act over screen images to accomplish tasks~\citep{shaw2023pixels, gou2024navigating}. These agents typically rely on supervised learning or modular grounding to interpret screen content and follow instructions.

In parallel, research on web agents, e.g., WebDreamer~\citep{gu2024your}, has demonstrated the value of simulating possible action outcomes before execution using learned world models. While these works focus on web environments composed of HTML documents and semantic DOM elements, mobile UI agents must operate on image-based environments where no structured metadata is available, and where understanding visual context (e.g., button layouts, app themes) is crucial. Moreover, many existing web agents rely on large-scale language models to reason about structured actions (e.g., clicking an HTML tag) and cannot be directly applied to purely visual screen-based interactions.

\ourmethod\ complements and extends this line of work by introducing a visually grounded UI simulation framework. It enables agents to predict and visualize the outcome of an action before executing it. This simulation capability not only enhances UI testing and debugging but also unlocks forward-planning capabilities for agents in vision-only environments, facilitating tasks like multi-step navigation or goal-directed interaction in mobile apps. In doing so, \ourmethod\ serves as a general-purpose platform for developing and training UI agents, analogous to how world models have enabled model-based planning in web-based agent research.

\section{\ourmethod}

\begin{figure*}[t]
  \includegraphics[width=\linewidth]{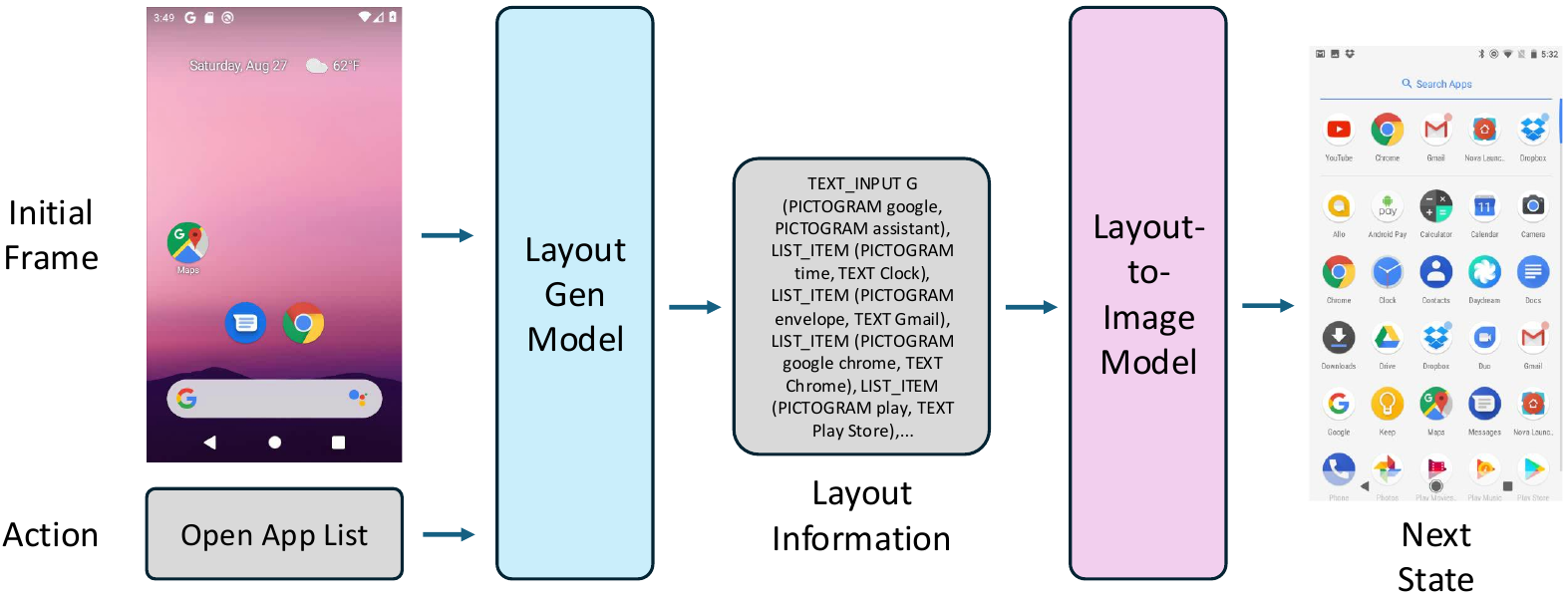} 
  \caption {Overview of \ourmethod's two-stage generation pipeline. Given an initial UI frame and a user action (e.g., "Open App List"), the Layout Generation Model predicts a structured layout description of the next UI state. This layout information—expressed in terms of UI components and their semantics—is then passed to a Layout-to-Image model, which synthesizes a realistic UI screen representing the predicted next state.}
\label{fig:method}
\end{figure*}

Our proposed UI simulator, \ourmethod, is designed to dynamically simulate mobile user interface environments directly from visual input, fundamentally addressing the limitations of static analysis and resource-intensive, device-based testing. As shown by Figure \ref{fig:method}, \ourmethod\ operates on a novel two-stage pipeline. It firstly generates layout information based on the input screen image and user action, then generates the next screen image based on the layout information. This enables both fine-grained control over UI transitions and the generation of high-fidelity, visually consistent screen images.

The benefits of this two-stage pipeline are substantial. By explicitly predicting the layout before rendering the UI image, our system gains fine-grained control over both the structure and content of the generated screen. This decoupled design breaks down the complex task of image-to-image UI transformation into more manageable sub-problems, mirroring the Chain-of-Thought~\citep{wei2022chain} method in large language models (LLMs), where reasoning is made more effective by decomposing complex tasks into intermediate steps. In contrast, end-to-end models must implicitly infer the target layout while synthesizing pixels, which often leads to degraded structure, lower coherence, and reduced controllability.

Our approach also aligns with recent advancements in large-scale image generation models, such as DALL-E 3~\citep{betker2023improving}, where providing more detailed and structured prompts or conditioning information leads to significantly better generative performance and adherence to user intent. Similarly, our second stage greatly benefits from a rich, explicit layout prompt, resulting in more accurate, coherent, and visually plausible UI images. Furthermore, the mapping between a well-defined UI layout and its visual representation is often close to one-to-one, much like specific HTML code predictably generates a particular website. This inherent consistency ensures that the generated UI remains visually coherent with the original phone system style and overall aesthetic, minimizing information loss and preserving fidelity across dynamic transitions.

\subsection{Layout Information Generation}\label{sec:layout-pred} The first stage of our pipeline, Layout Information Generation, is responsible for interpreting a user action on an initial UI screen and translating it into a structured, machine-readable representation of the predicted next UI state. The inputs to this stage are the initial phone screen image and a textual description of the user action (e.g., "open email app", "click search box", "scroll down").

We finetune an open-source vision language model (VLM) Qwen2-VL-7B-Instruct~\citep{wang2024qwen2} for layout information generation. The model is trained to understand the visual context of the UI and the semantic meaning of the user action, predicting how the screen layout will change. The training data is created using Android in the Wild dataset~\citep{rawles2023androidinthewild}, a rich collection of real-world mobile interactions. We extracted frame pairs from the trajectories in the dataset, ensuring diversity in applications and interaction types. For each pair, we annotated the user action that occurred between the initial and subsequent frame using Gemini 1.5 Pro~\citep{team2023gemini}, capturing the intent and locus of the interaction. Besides, we use ScreenAI~\cite{screenai}, a VLM for UI understanding, to systematically annotate the second frame, extracting its complete structural and semantic information. This annotation includes details such as UI element names, element descriptions, and bounding box coordinates. Finally, each training example consists of the initial screen frame, the user action annotated by Gemini, and the layout information of the subsequent frame. After finetuning on the constructed data, the model is able to effectively transform a visual input and an action into a precise, structured UI blueprint.

By decomposing the complex task of UI transition prediction and introducing an explicit layout information generation stage, \ourmethod\ delegates the reasoning process to a VLM. This approach leverages the capability of VLM to interpret multimodal inputs—an initial UI screen image and a textual user action—and plan the subsequent structural changes to the UI. Recent studies, such as DeepSeek-R1~\citep{guo2025deepseek}, have demonstrated that LLMs are highly capable in reasoning-intensive tasks. In contrast, most existing end-to-end image and video generation models typically follow a one-to-one paradigm, directly mapping prompts to outputs without incorporating an intermediate reasoning step. However, UI transition prediction inherently requires a deeper form of reasoning about user intent and how it will manifest in logical and structural UI changes. By offloading this high-level decision-making to a VLM, our approach effectively separates it from low-level pixel rendering, leading to more robust generalization, enhanced structural fidelity, and semantically coherent generation of subsequent UI states.

\subsection{Layout-to-Image Generation}
The second stage takes the abstract layout predicted in Stage 1 and renders it into a high-fidelity, pixel-level UI image.  We use an image diffusion model~\citep{ho2020denoising} that is specifically trained to synthesize visually compelling phone screen images from structured UI layout descriptions. The model was pretrained on a large-scale collection of diverse UI layout-image pairs, utilizing the same UI layout annotation system mentioned in Section~\ref{sec:layout-pred} to derive the textual layout inputs. By conditioning on the rich and structured layout information, the model can generate visually realistic and coherent UI screens that accurately reflect the intended structural and stylistic changes. As shown by advanced image generation models like DALL-E 3~\citep{betker2023improving}, more detailed prompts lead to significantly better generative outcomes. our two-stage pipeline provides the diffusion model with a highly refined prompt, i.e., layout information, which is a key reason \ourmethod outperform end-to-end baselines that attempt to generate the next screen directly from the initial image and action.

Another advantage of our architecture lies in its data scalability for pretraining. Collecting large-scale, high-quality layout-image pairs, e.g., crawling web UIs and annotating them, is relatively straightforward and can be automated at scale. In contrast, acquiring clean image-action-image triplets for end-to-end UI transition modeling is significantly more challenging. Screen operations recordings crawled from the web are often noisy and visually cluttered, while manually scaling up phone screen recordings is expensive and time-consuming. Our design allows us to easily leverage powerful diffusion models pretrained on readily available layout-image data, bypassing the bottlenecks associated with scarce dynamic interaction data.

\section{Experiments}
We detail our experimental setup in Section~\ref{sec:expr-setup}, including data construction, model training, and the baselines we used. We then show experimental results in Section~\ref{sec:res}.

\subsection{Experimental Setup}\label{sec:expr-setup}

\paragraph{Data Preparation.} We construct our dataset from the Android in the Wild~\citep{rawles2023androidinthewild}. In this dataset, each trajectory represents the completion of a high-level user goal (e.g., ``Find the cheapest hotel in Austin''). Within these goal-oriented trajectories, specific frames are manually tagged as keypoints, marking intermediate states towards the overall objective, e.g., the moment an app is opened, text is typed into a search box, or a results page is displayed.

We constructed our dataset of UI state transitions by leveraging the tagged keypoints in the original dataset. Specifically, for any two consecutive keypoints in a trajectory, we annotate the user action between them with Gemini, and the layout information for the second frame with an UI layout annotation system, as described in Section 3.1. Finally, we get 28306 examples in total. We subsample 27306 examples for training, with the remaining trajectories reserved for evaluation.

% \textcolor{red}{maria video 28306}
% \textcolor{red}{aitw single-step-49 28453}

\paragraph{Model Training.}
For the first stage of our pipeline, we finetuned a Qwen2-VL-7B-Instruct on our constructed dataset of initial UI images, user actions, and corresponding next-state layout annotations. We used LoRA~\citep{hu2022lora} for training efficiency. We used a learning rate of 1e-4, LoRA rank 4, and LoRA alpha 4. We trained the model for 5 epochs with batch size of 32. For the second stage, we leverage a pretrained layout-to-image diffusion model. The model was pretrained on large-scale collections of UI layout-image pairs, and can robustly generate high-fidelity UI screens from structured layout inputs.

\paragraph{Baselines.} We compare \ourmethod\  against two end-to-end generative baselines. The first baseline CogV-Image is an end-to-end image generation model, takes the initial UI screen image and the textual user action as direct inputs and generates the next UI screen image in a single step. Since UI interactions are inherently temporal, we also develop the second baseline CogV-Video, which is an end-to-end video generation model. It also takes the initial UI screen image and user action as input, but it is trained to generate a short video sequence depicting the UI transition. For comparison with our single-frame output, the final frame of the generated video sequence is used as the predicted next UI screen. Both of the two models are based on CogVideoX-5b-I2V~\citep{yang2024cogvideox} and finetuned on the same training data for \ourmethod.

\paragraph{Evaluation Metric.} We evaluate baselines and \ourmethod\ with Fréchet Inception Distance (FID)~\citep{heusel2017gans}, which is a widely recognized metric for assessing the quality and diversity of images generated by generative models. 

\begin{table}[t] % 'htbp' is a placement specifier: here, top, bottom, page
\centering % To center the table on the page
 % Add a descriptive caption
 \caption{Comparison of FID scores between our method (UISim) and two end-to-end baselines. Lower FID indicates higher visual fidelity. UISim significantly outperforms both baselines, demonstrating the effectiveness of our two-stage layout-guided generation approach in producing realistic and coherent UI screens.}
\label{tab:res} % Add a label for cross-referencing
\begin{tabular}{lr} % 'l' for left-aligned text column, 'r' for right-aligned number column
\toprule % Top rule from booktabs
Model & FID \\ % Column headers
\midrule % Middle rule from booktabs
CogV-Image & 98.37 \\
CogV-Video & 99.51 \\
Ours & \textbf{61.64} \\
\bottomrule % Bottom rule from booktabs
\end{tabular}
\end{table}

\subsection{Experimental Results}\label{sec:res} Table \ref{tab:res} shows the experimental results, demonstrating that \ourmethod\ consistently outperforms both the end-to-end image generation and end-to-end video generation baselines. This superior performance validates the effectiveness of our two-stage architecture, confirming that decoupling layout prediction from image synthesis leads to significantly higher fidelity and more realistic UI simulations. The explicit intermediate layout representation allows our model to maintain visual consistency and generate coherent UI states more effectively than approaches that directly generate pixels from initial image-action pairs.

\section{Conclusion}
We introduced \ourmethod, a novel two-stage image-based simulator for mobile user interfaces that enables realistic and controllable UI state transitions from visual inputs. Our method leverages a vision-language layout prediction model followed by a layout-conditioned image generation model. This design provides strong structural control and high visual fidelity, addressing the limitations of prior end-to-end UI generation methods.

Our system is trained on a rich dataset of real-world UI transitions with annotated user actions and layouts. Experimental results demonstrate that \ourmethod\ significantly outperforms competitive baselines—achieving a 36.73 improvement in Fréchet Inception Distance—while producing more coherent and semantically meaningful UI sequences.

Beyond UI prototyping and testing, \ourmethod\ opens up new possibilities for interactive agent training and visual planning in screen-based environments. Future work includes integrating multi-step simulation, expanding to multimodal UI representations, and exploring tighter loops between layout reasoning and generative feedback. We believe \ourmethod\ sets a foundation for more scalable, flexible, and intelligent interfaces in the next generation of human-computer interaction systems.

\newpage
\bibliography{colm2024_conference}
\bibliographystyle{colm2024_conference}

%%%%%%%%%%%%%%%%%%%%%%%%%%%%%%%%%%%%%%%%%%%%%%%%%%%%%%%%%%%%

\end{document}